\pdfoutput=1

\documentclass[11pt]{article}

\usepackage{acl}

\usepackage{times}
\usepackage{latexsym}
\usepackage{CJKutf8}
\usepackage{xspace}
\usepackage{tcolorbox}
\usepackage{pinyin}
\usepackage{makecell}
\usepackage{booktabs}
\usepackage{rotating}
\usepackage{multirow}

\usepackage[T1]{fontenc}

\usepackage[utf8]{inputenc}

\usepackage{microtype}

\newcommand{\qtuna}{\textsc{qtuna}\xspace}
\newcommand{\mqtuna}{\textsc{mqtuna}\xspace}

\title{Understanding the Use of Quantifiers in Mandarin}

\author{Guanyi Chen \and Kees van Deemter \\
  Department of Information and Computing Sciences \\
  Utrecht University \\
  \texttt{\{g.chen, c.j.vandeemter\}@uu.nl}} 

\begin{document}
\maketitle
\begin{abstract}
We introduce a corpus of short texts in Mandarin, in which quantified expressions figure prominently. We illustrate the significance of the corpus by examining the hypothesis (known as Huang's ``coolness'' hypothesis) that speakers of East Asian Languages tend to speak more briefly but less informatively than, for example, speakers of West-European languages. The corpus results from an elicitation experiment in which participants were asked to describe abstract visual scenes. We compare the resulting corpus, called \mqtuna, with an English corpus that was collected using the same experimental paradigm. The comparison reveals that some, though not all, aspects of quantifier use support the above-mentioned hypothesis. Implications of these findings for the generation of quantified noun phrases are discussed. \mqtuna is available at: \url{https://github.com/a-quei/qtuna}.
\end{abstract}
\begin{CJK}{UTF8}{gbsn}
\section{Introduction}





Speakers trade-off clarity against brevity \citep{grice1975logic}. It is often thought that speakers of East Asian languages handle this trade-off differently than those who speak Western European languages such as English~\citep{newnham1971about}.
This idea was elaborated in~\citet{huang1984distribution}, when Huang borrowed a term from media studies, hypothesizing that Mandarin is ``cooler'' than English in that the intended meaning of Mandarin utterances depends more on context than that of their English counterparts; in other words, Mandarin speakers make their utterances shorter but less clear than English speakers. This ``coolness" hypothesis is often worded imprecisely, conflating (a) matters that are built into the grammar of a language (e.g., whether it permits {\em number} to be left unspecified in a given sentence position), and (b) choices that speakers make from among the options that the grammar permits. Here we focus on the latter.

Studies of coolness have often focused on referring expressions (e.g., \citet{van-deemter-etal-2017-investigating, chen-etal-2018-modelling, chen-van-deemter-2020-lessons, chen2022computational}). The present paper focuses on {\em quantification}, as in the Quantified Expressions (QEs) ``\emph{All A are B}'', ``\emph{Most A are B}'', and so on. 
In a nutshell, we want to know whether Mandarin speakers use QEs less clearly, and more briefly, than English ones.

We report on an elicitation experiment, \mqtuna, inspired by the \qtuna experiment of ~\citet[][see \S\ref{sec:qtuna}]{chen-etal-2019-qtuna}. The experiment asks Mandarin speakers to produce sequences of QEs to describe abstract visual scenes. Sequences of QEs that are used to describe visual scenes are called Quantified Descriptions~\citep[QDs,][]{chen-etal-2019-qtuna}.
The \mqtuna corpus will enable researchers to investigate a wide range of questions about quantification in Mandarin. We illustrate this potential by comparing the corpus with the English \qtuna corpus from the perspective of coolness and we ask how our findings impact computational models of the production of QDs.

In sum, our contribution is two-fold: 
\begin{enumerate}
    \item We constructed, annotated and analysed the \mqtuna corpus;
    \item We compared \mqtuna to \qtuna from the perspective of \citeauthor{huang1984distribution}'s Coolness hypothesis.
\end{enumerate}
\section{\qtuna Experiment} \label{sec:qtuna}

\begin{figure}
    \centering
    \frame{\includegraphics[scale=0.3]{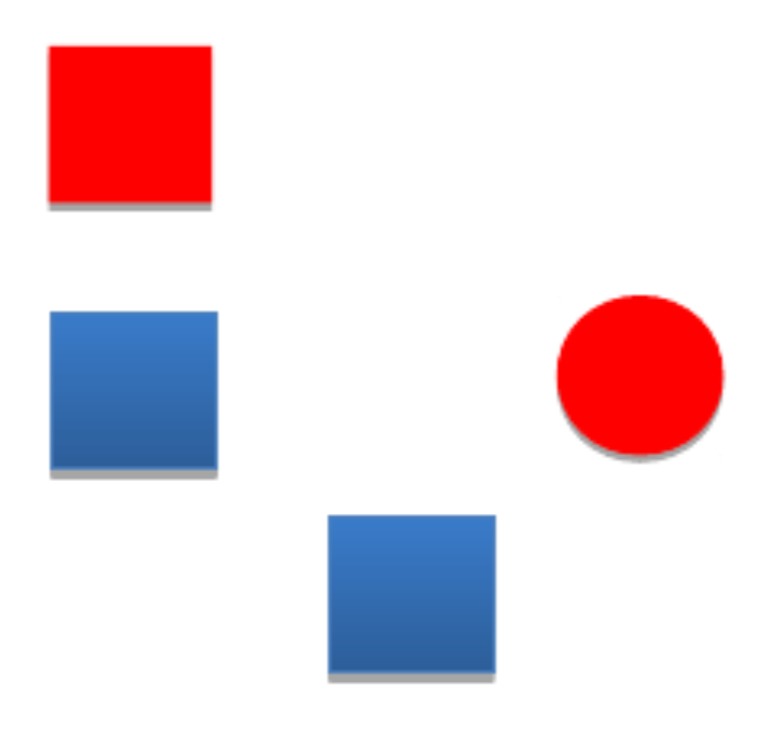}}
    \caption{An example scene from \qtuna.}
    \label{fig:qtuna}
\end{figure}

A growing body of empirical work has studied how people understand and produce quantifiers~\citep{moxey1993communicating, szymanik2010comprehension,grefenstette2013towards,herbelot2015building,sorodoc2016look}. These studies have focused on a limited number of quantifiers (chiefly ``\emph{all}'', ``\emph{most}'', ``\emph{many}'', and ``\emph{no}'').

In Natural Language Generation (NLG), the \qtuna corpus was built to study how English speakers use QDs to describe a visual scene (Figure~\ref{fig:qtuna}). 
Participants were free to (1) describe a visual scene in whatever way they want, (2) use as many sentences as they choose, and (3) use any sentence pattern that they choose. For example, for the scene in Figure~\ref{fig:qtuna}, a participant could say ``\emph{Half of the objects are blue squares. The other half are red objects. There is only one red circle.}''.
Given the domain contains four objects in no more than two shapes, this QD describes the scene completely and correctly.
Participants were told that their descriptions should allow readers to reconstruct the scene {\em modulo} location. Each scene contains $N$ objects (NB: $N$ is defined as \emph{domain size}), which is either a circle or a square and either blue or red. To test how domain size impacts the use of quantifiers, \qtuna experimented on 3 sizes, i.e., 4, 9, and 20.

Analysis of the resulting \qtuna corpus revealed that, as the domain size increase, (English) speakers (1) use more vague quantifiers (e.g., \emph{most} and \emph{few}); (2) use less complete QDs (NB: a QD is complete if the scene described is the only one {\em modulo} location 
that fits the description); (3) use more incorrect QDs (NB: a QD is incorrect if it is not true with respect to the scene); and (4) do not use longer QDs (measured in terms of the number of QEs).  
\section{Research Questions} \label{sec:rq}

\paragraph{Are the \qtuna findings true for \mqtuna?} 

We are curious whether the above-mentioned findings about \qtuna (see \S\ref{sec:qtuna}) hold true for \mqtuna. We expected that domain size affects speakers of different languages in the same way, so these findings should hold for both corpora in the same way.

\paragraph{Are Mandarin QDs briefer and less clear than English QDs?}

``Coolness'' says Mandarin speakers speak more briefly and less clearly than English speakers.
We check this hypothesis by comparing QDs in \qtuna and \mqtuna.

Regarding brevity, we are curious about the length of QDs. If Mandarin QDs are briefer than English QDs, then we expect QDs in \mqtuna to contain less QEs than those in \qtuna. 

Regarding clarity, 
if Mandarin speakers utter QDs in a less clear way, we expect to see more vague quantifiers in \mqtuna than in \qtuna and, more importantly, fewer logically complete QDs.
\section{\mqtuna Experiment}

We followed the same methodology as in the \qtuna experiment, re-using scenes of the \qtuna experiment, inheriting its experimental design, and translating its instructions participants.

\subsection{Materials}

To prepare materials for the \mqtuna experiment, we sampled scenes from \qtuna following two steps. First we eliminated all scenes all of whose objects share the same properties. For instance, we removed all scenes that can be described completely by a single QD like ``\emph{all objects are red circles}''.
Next, for each domain size (i.e., 4, 9, or 20), we randomly sampled 5 scenes from \qtuna.
In the second step, to familiarise participants with the experiment, we added a practice situation that uses a $N=4$ scene whose objects are the same. 

For the instructions, we translated the instructions of \qtuna 
(Appendix~\ref{sec:instruction}).
More specifically, the instruction told subjects that (1) they should finish the experiment in limited time (i.e., 20 minutes); (2) their descriptions would then be used in a reader experiment where readers are asked to re-construct the scenes; (3) they should not enumerate and not say where in the grid a particular object is located.

\subsection{Design, Participants, and Procedure}

Data from 31 participants were collected for domain sizes $N=4$, $9$ and $20$ ($N$ is the number of objects in the scene). See Appendix~\ref{sec:participants} for details about participants.
Participants were asked to read the instruction first and to complete the experiment (16 situations) in one sitting. 

\subsection{The \mqtuna Corpus} \label{sec:mqtuna}

\begin{table*}[t]
    \centering
    \footnotesize
    \begin{tabular}{lp{14cm}}
    \toprule
       $N$ & Description \\ 
    \midrule
        4 & \makecell[l]{所有都是蓝色，方块是圆形三倍。$|$ \emph{All objects are blue. The number of squares is triple that of circles.}}\\
        4 & \makecell[l]{所有图形都是蓝色的。但是只有一个圆。$|$ \emph{All objects are blue but there is only one circle.}}\\ \midrule
        9 & \makecell[l]{所有的圆圈是红色的。方块都是蓝色的。方块的数量少于圆圈的数量。\\\emph{All circles are red. All squares are blue. There are fewer squares than circles.}}\\ 
        9 & \makecell[l]{方块是圆圈数量的三倍。全部为红色。$|$ \emph{The number of squares is triple that of circles. All of them are red.}}\\ \midrule
        20 & \makecell[l]{图中红色蓝色方块圆球数量相差不大。$|$ \emph{There is no big difference between the numbers of all combinations.}}\\
        20 & \makecell[l]{一半红色，一半蓝色。红色方块比蓝色方块多。蓝色圆圈多于红色圆圈。\\\emph{Half of the objects are red, the other half of them are blue. There are more red squares than blue squares}\\ \emph{and more blue circles than red circles.}}\\
    \bottomrule
    \end{tabular}
    \caption{List of example descriptions from the \mqtuna corpus, with their annotations. $N$ indicates domain size.}
    \label{tab:mqtuna_example}
\end{table*}

The resulting \mqtuna corpus contains 465 valid QDs and 1175 QEs. There are 155 QDs for each domain size and there are 383, 386, and 406 QEs for $N=4$, $N=9$, and $N=20$ respectively.
Table~\ref{tab:mqtuna_example} lists a number of examples QDs in \mqtuna.

We annotated the use of quantifiers in \mqtuna, viewing quantifiers that have the same meaning (e.g., ``所有'' (``suoyou", all) and ``全部'' (``quanbu", all) as identical.
See Appendix~\ref{sec:quantifiers} for a list of top-10 quantifiers and their usage in \mqtuna.

As for quantifier use, the quantifier ``所有'' (\suo3\you3; \emph{all}) and ``一半'' (\yi2\ban4; \emph{half}) are two of the most frequent quantifiers. 
In the top-10 most frequent quantifiers of \mqtuna, 4 are vague, including ``绝大多数'' (\emph{overwhelming majority}), ``大多数'' (\emph{most}), ``多数'' (\emph{most}), ``少数'' (\emph{minority}). 
This is very different from \qtuna, where only 1 vague quantifier (i.e., \emph{most}) is in top-10. 
Appendix~\ref{sec:quantifiers} also presents lists of crisp and vague quantifiers.
\section{Analysis} \label{sec:analysis}

Focusing on the research questions of \S\ref{sec:rq}, we analyse the \mqtuna corpus (\S\ref{sec:analysis_mqtuna}), and we compare \mqtuna with \qtuna (\S\ref{sec:compare_mqtuna_qtuna}). We conclude with a few post-hoc observations (\S\ref{sec:observation}).

\subsection{Analysing \mqtuna} \label{sec:analysis_mqtuna}

\begin{table}[t]
    \centering
    \small
    \begin{tabular}{lccc}
        \toprule
         & $N=4$ & $N=9$ & $N=20$\\ \midrule
         Quantified Description & 155 & 155 & 155 \\
         Quantified Expression & 383 & 386 & 406 \\ \midrule
         Complete Description & 122 & 19 & 5 \\
         Incomplete Description & 33 & 136 & 150 \\
         Vague Quantifier & 25 & 143 & 184 \\
         Wrong Description & 7 & 14 & 30 \\
         \bottomrule
    \end{tabular}
    \caption{Frequencies of major QE types in the different subcorpora of \mqtuna. 
    }
    \label{tab:mqtuna_analysis}
\end{table}

To check whether the findings of \qtuna 
(\S\ref{sec:qtuna}) hold for \mqtuna, we annotated each QD with whether it is logically complete and whether it is correct with respect to the corresponding scene; we also annotated each QE with whether it uses a vague quantifier or not. Annotation details can be found in Appendix~\ref{sec:annotation}.
To avoid compromising the comparison between 
\mqtuna and \qtuna, we did not only annotate \mqtuna but we also re-annotated the \qtuna corpus, using the same annotators following the same set of principles.
Table~\ref{tab:mqtuna_analysis} charts the results.

\paragraph{Vagueness.} We identified 57, 201, and 234 QEs that contain vague quantifiers out of 383, 386, and 406 QEs from the three sub-corpora, confirming that vagueness is more frequent with increasing domain size. 
This was confirmed by a binary logistic regression test ($p < .0001$, adjusted $p < .0001$\footnote{The p-value was adjusted by Bonferroni correction}).

\paragraph{Completeness.} We observed 33, 136, and 150 logically incomplete QDs from the three sub-corpora. A binary logic regression test confirms that there are more logically incomplete QDs in larger domains ($p < .0001$, adjusted $p < .0001$).

\paragraph{Correctness.} The 3 subcorpora contained 7, 14, and 30 wrong QDs, so more incorrect QDs are used in larger domains ($p < .0001$, adjusted $p < .0001$) using a binary logic regression test.

\paragraph{Length.} QDs in larger domains in \mqtuna contain more QEs than those in smaller domains. We computed the Pearson correlation between the domain size and the QD length. After Bonferroni correction, the difference fell just short of significance ($p = 0.1025$, adjusted $p=0.615$). 

In a nutshell, all findings of \qtuna are also true for \mqtuna.

\subsection{\mqtuna and \qtuna: Initial Comparison} \label{sec:compare_mqtuna_qtuna}

To compare QDs in Mandarin and English, we considered all the scenes in \mqtuna  
and re-annotated them using the same annotators 
in \S\ref{sec:analysis_mqtuna}.

\paragraph{Brevity.} We compared the length of QDs in \qtuna and \mqtuna and found that QDs in \mqtuna are longer than those in \qtuna in every sub-corpus. This rejects our hypothesis that Mandarin speakers prefer brevity and, thus, produce shorter QDs than English speakers.

\begin{table}[t]
    \centering
    \small
    \begin{tabular}{cccccc}
        \toprule
         & \multicolumn{2}{c}{\textsc{qtuna}} & \multicolumn{2}{c}{\textsc{mqtuna}} & \\ \cmidrule(lr){2-3} \cmidrule(lr){4-5} 
         $N$ & C & I & C & I & p-value \\ \midrule
         4 & 298 & 32 & 122 & 33 & $p<.001$ \\
         9 & 77 & 113 & 19 & 136 & $p<.0001$ \\
         20 & 4 & 186 & 5 & 155 & $p=.5$ \\
         all & 379 & 331 & 146 & 319 & $p<.0001$ \\
         \bottomrule
    \end{tabular}
    \caption{Numbers of complete (C) and incomplete (I) QEs in \textsc{qtuna} and \textsc{mqtuna}. $N$ is domain size.}
    \label{tab:compare_complete}
\end{table}

\paragraph{Completeness.} Table~\ref{tab:compare_complete} reports the number of logically complete QDs in \qtuna and \mqtuna, respectively. 379 out of 710 QDs in \qtuna are logically complete while 146 out of 465 QDs in \mqtuna are complete. Using a Chi-squared test, this confirms that there are more complete QDs in \qtuna than in \mqtuna ($\PYchi^2(2, N=1175)=54.93, p<.0001$, adjusted $p<.0001$). 
Mandarin speakers produce longer but less logically complete QDs. Interestingly, if we look into more details (see Table~\ref{tab:compare_complete}), the difference only exists in domain sizes 4 and 9. 
We suspect that both English and Mandarin speakers find it hard to come up with a logically complete QD if the domain size is large.

\paragraph{Vagueness.}

In \qtuna, 222 of the 1342 QEs were vague whereas, in \mqtuna, 352 of the 1175 QEs were vague. 
A Chi-squared test confirms that Mandarin speakers used more vague quantifiers than English speakers ($\PYchi^2(2, N=2517)=64.04, p<.0001$, adjusted $p<.0001$).

\subsection{Post-hoc Observations} \label{sec:observation}

\paragraph{Surface Forms.} We observed that QEs in \mqtuna are generally realised in three kinds of forms: (1) ``Q A 是 B'' (``\emph{Q A are B}''), where ``Q'' is a quantifier, for example, ``大部分 A 是 B'' (``\emph{most A are B}''); (2) ``A 中 Q 是 B'' (``in A, Q are B''); and（3）``B 在 A 中 占 Q'' (``\emph{B takes up Q of A}'').

\paragraph{A-Drop.} Akin to the previous findings that pronouns and nouns are often dropped in Mandarin NPs~\citep{huang1984distribution, osborne2015survey}, we found that nouns that take up A positions in the above forms are also often dropped (henceforth, \emph{A-drop}), for example, saying ``B 占 Q'' (``\emph{B takes up Q}''). In \mqtuna, we found 304 out of 1175 QEs (approximately 25.87\%).

\paragraph{Plurality.} \citet{van1998adverbial} pointed out that Mandarin briefer in that plurality is often not expressed explicitly. Consistent with this, we found that in \mqtuna, numbers are 
rare.
This makes a QE in Mandarin sometimes less informative than an English QE, Mandarin QDs are less likely 
to be logical complete. For example, Mandarin QE ``图片中有红色方块'' could mean ``there are red squares'' or ``there is a red square''.
\section{Discussion}

We have presented and analysed the \mqtuna corpus of quantifier use in Mandarin. 

\paragraph{Coolness.} We assessed the coolness hypothesis by analysing \mqtuna and comparing \qtuna and \mqtuna. 
As for the brevity of QDs, we found both evidence (i.e., Mandarin speakers often performed A-drop and expressed plurality implicitly) and counter-evidence (i.e., Mandarin speakers uttered longer QDs than English speakers).

As for the clarity of QDs, we confirmed that the Mandarin corpus (\mqtuna) contains significantly more {\em incomplete} QDs and {\em vague} quantifiers than its English counterpart (\qtuna).

\paragraph{Generating QDs.} 
\citet{chen-etal-2019-generating} proposed algorithms for generating QDs (QDG algorithms). Let us list issues to be heeded when building QDG algorithms for Mandarin.

First, plurality plays an important role in the QDG Algorithms of \citet{chen-etal-2019-generating}. If these algorithms are to be adapted to Mandarin, then they should first ``decide" whether to realise the plurality of a QE explicitly, since this will influence how much information the QD should express in other ways. 
Second, modelling the meaning of vague 
quantifiers is vital for generating human-like QDs.
Since Mandarin speakers use vague quantifiers more frequently than English speakers, Mandarin QDG needs to handle a larger number of vague quantifiers and capture nuances between them, which is a difficult and data-intensive challenge.
Lastly, QD surface realisation in Mandarin needs to handle more syntactic variations than current QDG algorithms are capable of, 
because (1) a QE can be realised in multiple possible forms (see \S\ref{sec:observation}); (2) A-drop frequently happens; (3) Plurality can be expressed implicitly or explicitly.\\[1ex]
{\bf Future Work.} Our comparison between Mandarin and English was based on two corpora, \qtuna and \mqtuna, that were collected using elicitation experiments that were conducted following the same experimental paradigm, and using very similar sets of stimuli. Yet, {\em language} may not have been the only difference between these experiments; participants in \qtuna and \mqtuna are also likely to differ in terms of their {\em cultural background}, and possibly in terms of other variables, such as their education; 
There is no absolute guarantee that all our annotations are correct. To create an even playing field between the two corpora, we asked our annotators to re-annotate \qtuna. But although our annotator were native speakers of Chinese, they were merely fluent (not native) in English, which may have caused a difference in the way both corpora were annotated.
In future, it would be interesting to conduct even more tightly controlled experiments to tease apart the variable of language use from such possibly confounding variables.

Finally, our experiment has looked at a wide range of quantifiers. We also plan experiments that zoom in on specific subsets, such as the different ways in which variants of the quantifier ``\emph{most}'' can be expressed (cf., \citet{wang2007translating} and \S\ref{sec:mqtuna}).

\bibliography{anthology,custom}
\bibliographystyle{acl_natbib}

\appendix

\newpage
\section{Instruction} \label{sec:instruction}

\begin{figure}[ht]
    \begin{tcolorbox}[colback = white, boxrule = 0.3mm]
    \emph{\small
    您好，我们最近的研究关注于人描述物体集合的方法。为此，我们设计了一个小实验。在这个实验中，我们将给您展示一系列图片。在每张图片中，您将看到一定数量（16个）的图形。在看到每张图片后，我们需要您写一句或几句语法正确的中文句子。请注意：\\
    We’re interested in understanding how people describe sets of objects. To find out, we’re doing a small experiment, in which we’ll show you a number of situations in which some (16) objects are displayed on a screen. We’d like you to describe each situation in one or more grammatically correct Mandarin sentences. 
    \begin{itemize}
    \item[1] 您将在有限的时间（20分钟）内完成整个实验。 The experiment should take you less than 20 minutes.
    \item[2] 根据您写的描述，后续实验中的被试者会用它来在有限时间（总共20分钟）内重构图片。 “重构”的在这里仅表示图片中每种图形数量。因此在您的表述中，您不必描述每个图形在图片中的位置（例如：上方，在中间）。 Based on your description, a reader will try to “reconstruct” the situation. We use the word “reconstruct” loosely here, because the only thing that matters is the different types of objects that the sheet contains. Therefore, please do not say *where* in the grid a particular object is located (e.g., ”top left”, “in the middle”, “on the diagonal”).
    \item[3] 每个图形可能是方形也可能是圆形，可能是红色也可能是蓝色。后续负责重构的被试者也知晓这个信息。负责重构的被试者同时还知晓图片中图形的数量。这些信息都会被用在重构当中。Each object is a circle or a square, and either red or blue. Your reader knows this.
    \item[4] 请不要“枚举”图片中的图形，例如：图片中有一个红色的圆圈，两个蓝色的圆圈，和三个蓝色的方块。Please do not “enumerate” the different types of objects. For example, do not say “There is a red circle, two blue circles, and ...”.
    \end{itemize}
    以下是几个例子: \\
    Here are some Example:}
    (...)
    \end{tcolorbox}
    \caption{The sketch of the instruction of \mqtuna. }
\label{fig:q_mqtuna_instruction}
\end{figure}

\section{Participants} \label{sec:participants}

All of our participants are Mandarin native speakers.
21 subjects are undergraduate students in computer science from the Utrecht University. Each of the rest at least has a bachelor degree in any of computer science, statistics, and management. 11 subjects are female and 20 subjects are male.

\section{Quantifiers in \mqtuna} \label{sec:quantifiers}

Table~\ref{tab:mqtuna_quantifier} enumerates the top-10 quantifiers and their usage in \mqtuna. In what follows, we provide a list of vague quantifiers and a list of crisp quantifiers in \mqtuna.
\begin{itemize}
    \item Crisp Quantifiers: 所有 (\emph{all}), 只有 (\emph{only}), 比...多... (\emph{more}), 倍 (\emph{times}), 除了...都是... (\emph{all...except...}), 有 (\emph{there is}), 多于n倍 (\emph{more than n times}), 少于n倍 (\emph{less than n times}), 各半 (\emph{half...the other half...}), 相同 (\emph{same as}), 一半 (\emph{half}), 不同 (\emph{different amount of}), 一半以上 (\emph{more than half}), 没有 (\emph{no}), 少于 (\emph{less than}), 所有组合 (\emph{all possible combinations});
    \item Vague Quantifiers: 大部分 (\emph{most}), 小部分 (\emph{a small part of}), 绝大部分 (\emph{overwhelming majority}), 除了...大多数... (\emph{most...except...}), 少量的 (\emph{a few}), 远多于 (\emph{way more than}),  极少数 (\emph{a very few}), 多一点 (\emph{slightly more than}), 多不少 (\emph{greatly more than}), 相近 (\emph{close to each other}), 基本都是 (\emph{almost all}), 略少 (\emph{a bit less}), 略多 (\emph{a bit more}), 大约各半 (\emph{approximately half ... the rest ...}), 基本相同 (\emph{almost the same}), 多一些 (\emph{several more}), 多好几倍 (\emph{several times more}), 多得多 (\emph{much more}), n倍多一点 (\emph{slightly more than n times}), n倍少一点 (\emph{slightly 
   less than n times}), 大约一半 (\emph{approximately half}), 少数 (\emph{minority}).
\end{itemize}

\begin{sidewaystable*}
    \centering
    \small
    \begin{tabular}{lp{2cm}p{4.5cm}p{6cm}cccc}
    \toprule
        \multirow{2}{*}{Notation} & \multirow{2}{*}{English} & \multirow{2}{*}{Surface Form(s)} & \multirow{2}{*}{Example Quantified Expression(s)} & \multicolumn{4}{c}{Frequency} \\ \cline{5-8} 
         &&&& N=4  & N=9 & N=20 & Total\\ \midrule
        所有 & all & (所有)...都..., (全部)...都... & \makecell[l]{(全部)A都是B / A中(全部)都是B \\\emph{All A are B}} & 100 & 127 & 53 & 280 \\ 
        一半 & half & 一半，百分之五十 & \makecell[l]{一半A是B / A中的一半是B / B在A中占一半 \\\emph{Half A are B}} & 101 & 19 & 28 & 148 \\
        相同 & equal & 数量相同, 一样多, 个数一样 & \makecell[l]{A与B数量相同\\\emph{There is an equally number of A and B}} & 59 & 11 & 29 & 99 \\
        绝大多数 & overwhelming majority & 绝大部分, 绝大多数 & \makecell[l]{A中绝大多数是B / 绝大多数A是B / B在A中\\占绝大多数\\\emph{Almost A are B}} & 7 & 50 & 37 & 94 \\ 
        各半 & half ... rest ... & 各半, 一半一半, 一半...另一半... & \makecell[l]{BC在A中各半 / A中BC各半 / 一半的A是B, \\另一半是C\\\emph{Half of A are B, the other half of A are C}} & 60 & 6 & 24 & 90 \\ 
        比-多 & more & 比...多 & A比B多 & 10 & 28 & 48 & 96 \\
        大多数 & most & 大多数, 大部分 & \makecell[l]{A中大多数是B / 大多数A是B / B在A中占\\大多数\\\emph{Most A are B}} & 7 & 35 & 33 & 75 \\
        少数 & minority & 少数, 少部分 & \makecell[l]{A中少数是B / 少数A是B / B在A中占少数\\\emph{Minority of A are B}} & 5 & 31 & 24 & 60\\ 
        有 & exist & 有, 存在 & 图片中有A (\emph{There are A in the scene}) & 4 & 12 & 18 & 34 \\ 
        多数 & most & 多数 & \makecell[l]{A中多数是B / 多数A是B / B在A中占多数 \\ \emph{Most A are B}}  & 5 & 4 & 20 & 29\\
    \bottomrule
    \end{tabular}
    \caption{Top-10 most frequently occurring quantifiers with their English translation and Mandarin examples as well as frequencies in the three \mqtuna sub-corpora.} 
    \label{tab:mqtuna_quantifier}
\end{sidewaystable*}

\section{Annotating \mqtuna} \label{sec:annotation}

We asked our annotator to annotate logical completeness, correctness and vagueness based on the following principles:
\begin{enumerate}
    \item Logical Completeness: we asked our annotator whether s/he can fully recover the scene given a QD. For example, for a scene with 3 red circles and 1 blue square, one could say ``\emph{Most objects are red circles and there is only one blue square.}'' Though s/he uses a vague quantifier ``most'', we still can infer that, given domain size 4, ``most objects'' means 3 objects, and, therefore, this QD is logically complete. However, for a scene with 8 red circles and 1 blue circle, one could say ``\emph{All objects are circles and almost all of them are 8.}'' Though using ``almost all'' to describe ``8 out of 9'' is definitely correct, it does not necessarily mean ``8 out of 9'' but possibly mean ``7 out of 9''. Therefore, this QD is not logically complete;
    \item Correctness: we asked our annotator to annotate a QD as ``incorrect'' if and only if the QD contains definitely incorrect information, for example, saying a ``red object'' blue or describing a scene with 3 red squares and 1 blue square as ``\emph{half of the objects are red}'';
    \item Vagueness: our annotator decided whether a QE uses a vague quantifier based on the vague quantifier list in Appendix~\ref{sec:quantifiers}.
\end{enumerate}

\end{CJK}
\end{document}